\documentclass[11pt,a4paper]{article}
\usepackage[hyperref]{eacl2021}
\usepackage{times}
\usepackage[T1]{fontenc}
\usepackage{latexsym}

\usepackage{graphicx}
\usepackage{multirow}
\usepackage{float}
\usepackage{siunitx}
\graphicspath{ {./images/} }

\usepackage{microtype}

\aclfinalcopy 

\title{Adapting MARBERT for Improved Arabic Dialect Identification:\\ Submission to the NADI 2021 Shared Task}

\author{Badr AlKhamissi\thanks{\,\, Equal contribution.} \\
        Independent \\
        \texttt{\small badr [at] khamissi.com} \\
\And
    Mohamed Gabr\footnotemark[1] \\
        Microsoft EGDC \\
        \texttt{\small mohamed.gabr [at] microsoft.com} \\
\AND 
    Muhammed ElNokrashy \\
        Microsoft EGDC \\
\And
    Khaled Essam \\
        Mendel.ai \\ 
}

\date{}

\begin{document}
\maketitle

\begin{abstract}
In this paper, we tackle the Nuanced Arabic Dialect Identification (NADI) shared task \citep{mageed2021nadi} and demonstrate state-of-the-art results on all of its four subtasks. Tasks are to identify the geographic origin of short Dialectal (DA) and Modern Standard Arabic (MSA) utterances at the levels of both country and province. Our final model is an ensemble of variants built on top of MARBERT that achieves an F1-score of $34.03\%$ for DA at the country-level development set---an improvement of $7.63\%$ from previous work.

\end{abstract}

\section{Introduction}




The Arab World is a vast geographical region that covers North Africa and Southwest Asia, boasting a population of around 400M that speak different derivatives of a common language. However, by virtue of its size and cultural variety, there exists a dialect continuum across the region wherein the language varieties of neighboring peoples may differ slightly, but distant regions can become mutually unintelligible. This continuum is referred to as Dialectal Arabic (DA) and is the ``Low'' variety of modern Arabic \emph{diglossia}. On the other hand, the ``High'' variety is referred to as Modern Standard Arabic (MSA) and is used in formal settings such as academia, mass media, and legislation---and is taught through the formal education system in most Arab countries. This standard variant emerged gradually, but most notably with the advent of the printing press in the 19th century. It diverged from Classical Arabic (CA) into a more simple version that is now used across the Arab World.

The modern vernacular dialects (DA) differ along several dimensions, including pronunciation, syntax, morphology, vocabulary, and even orthography. Dialects may be heavily influenced by previously dominant local languages. For example, Egyptian variants are influenced by the Coptic language, while Sudanese variants are influenced by the Nubian language. 

In this paper, we study the classification of such variants and describe our model that achieves state-of-the-art results on all of the four Nuanced Arabic Dialect Identification (NADI) subtasks \citep{mageed2021nadi}. The task focuses on distinguishing both MSA and DA by their geographical origin at both the country and province levels. The data is a collection of tweets covering $100$ provinces from $21$ Arab countries. The code has been made open-source and available on GitHub\footnote{\url{https://github.com/mohamedgabr96/NeuralDialectDetector}}. 



\section{Related Work}

The first efforts to collect and label dialectal Arabic data goes back to 1997 \citep{callhomedataset}. However, studying DA in NLP started to gain traction in recent years as more digital Arabic data became available, especially with the rise of online chatting platforms that place no restrictions on the syntax, style, or formality of the writing.
\citet{AOC:1} labelled the Arabic Online Commentary Dataset (AOC) through crowd-sourcing, then built a model to classify even more crawled data. \citet{abdelali2020arabic} provided the QADI dataset by automatically collecting dialectal Arabic tweets and labelling them based on descriptions of the author accounts, while trying to reduce the recall of written MSA and inappropriate tweets. \citet{abu-farha-magdy-2020-arabic} provided ArSarcasm: a dataset of labelled tweets. Originally for sarcasm detection, it also contains labels for sentiment and dialect detection.
Labelled dialectal Arabic challenges such as MADAR \citep{bouamor-etal-2019-madar} and NADI \citep{abdulmageed2020nadi, mageed2021nadi} (which started in 2020) shed light on the underlying challenges of the task. Both comprise labelled Arabic tweets but with class sets of different granularities. 


\citet{talafha2020multidialect} presents a solution that won the 2020 NADI shared task (Subtask 1.2) by continuing training AraBERT \citep{antoun2020arabert} using Masked Language Modelling (MLM) on 10M unlabelled tweets, then fine-tuning on the dialect identification task. On the other hand, the solution that won Subtask 2.2 uses a hierarchical classifier that takes as input a weighted combination of TF-IDF and AraBERT features to first classify the country, then invokes an ArabBERT-based, country-specific, province-level classifier to detect the province \cite{el-mekki-etal-2020-weighted}.

The large size of pretrained Transformer models hinders their applicability in many use cases. The usual number of parameters in such a model lies between 150M and 300M and needs between 500MB and 1GB of space to be stored. A novel technique to address these issues is proposed by \citet{houlsby2019parameterefficient}---bottleneck layers (``adapters'') are added to each transformer layer as an alternative to fine-tuning the entire pre-trained model when optimizing for downstream tasks \citep{houlsby2019parameterefficient, bapna-firat-2019-simple, pfeiffer2021adapterfusion}. Only these additional parameters (which can be 1\% or less of the size of the main model) need to be stored per downstream task, given that they are the only layers changing. Besides being light-weight and scalable, adapters offer several advantages over traditional approaches of transfer-learning: (1) They learn modular representations that are compatible with other layers of the transformer, (2) They avoid interfering with pre-trained knowledge, mitigating catastrophic forgetting and catastrophic inference---two common downsides of multi-task learning \citep{pfeiffer2020AdapterHub}.

\section{Datasets \& Subtasks}
\label{sec:data_tasks}

\begin{table}[h]
\centering
\begin{tabular}{lcc}
\hline
\textbf{Dataset} & \textbf{Country} & \textbf{Province} \\ \hline
\textbf{MSA} & Subtask 1.1      & Subtask 2.1       \\
\textbf{DA}  & Subtask 1.2      & Subtask 2.2       \\ \hline
\end{tabular}
\caption{Subtask ID per Dataset and Granularity Level}
\label{tab:subtasks}
\end{table}

\begin{figure}
    \centering
    \includegraphics[width=0.45\textwidth]{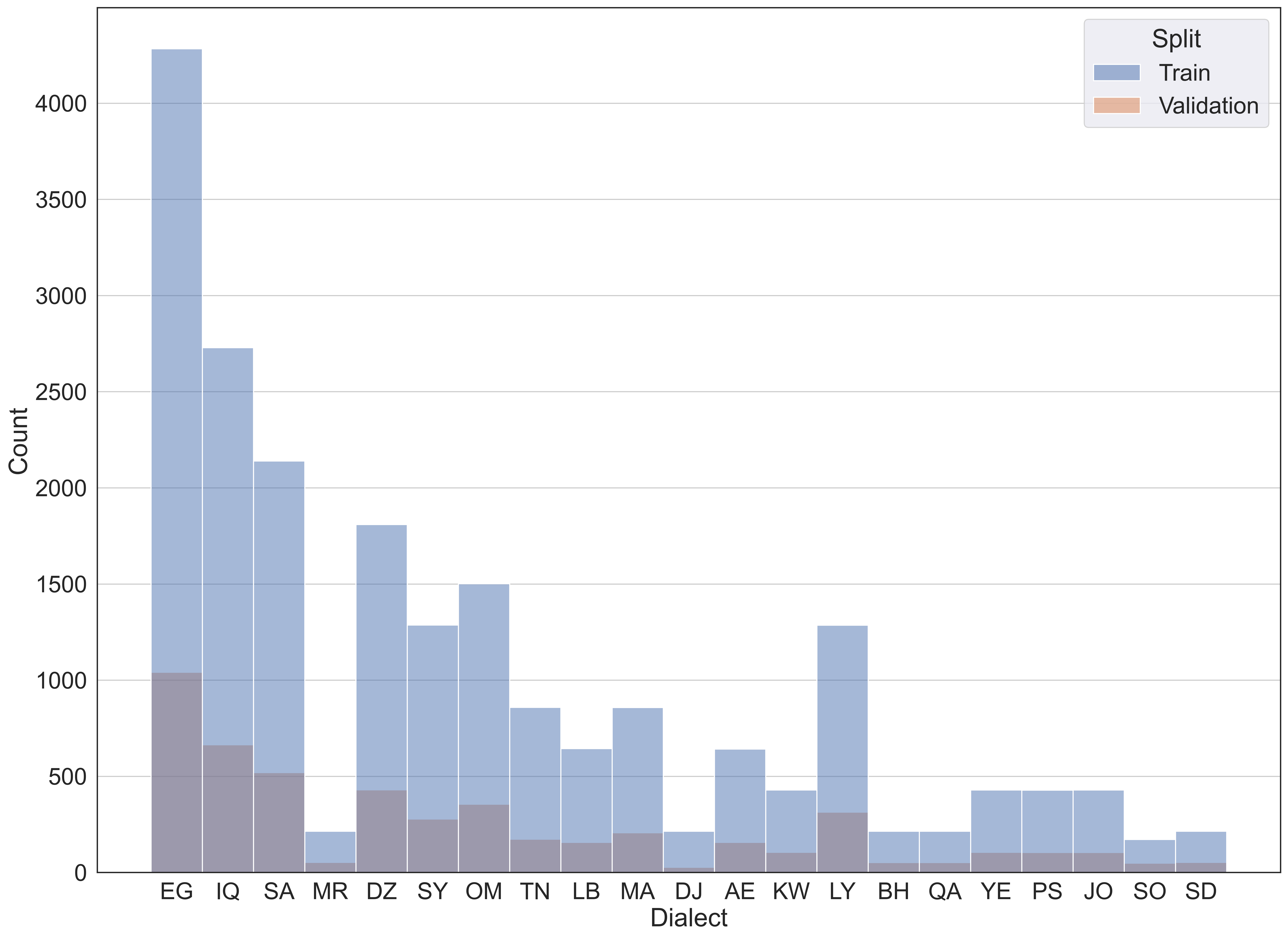}
    \caption{Train/Dev Corpora Sizes per Country (DA)}
    \label{fig:train_dev_counts}
\end{figure}

The second NADI shared task consists of four subtasks on two datasets (see Table \ref{tab:subtasks}). Each consists of $21$k tweets for training, $5$k for development and $5$k for testing collected from a disjoint set of users \citep{mageed2021nadi}. Both datasets are labelled at two levels of granularity: country, and province. The NADI task is the first to focus on sub-country level dialects. However, the data is extremely unbalanced, even at the country-level (see Figure \ref{fig:train_dev_counts}), with the most frequent class being Egypt (4283 instances), and the least common class being Somalia (172 instances). The data comes preprocessed with URLs replaced with the token `URL' and Twitter mentions replaced with the token `USER'. One of the main challenges of Arabic Dialect Identification is the high similarity of dialects in short utterances; many short phrases are commonly used in all dialects. Since the tweets are collected ``in the wild'' and DA is not formally defined (the same word can be written in a variety of ways), this makes it even more challenging to capture the meaning of a logical word unit across its different forms.


\section{System Description}
\label{sec:system}


Our model builds on MARBERT---a publicly available transformer model trained on $1$B multi-dialectal Arabic tweets \citep{mageed2020marbert}. It follows the BERT\textsubscript{BASE} architecture \citep{devlin-etal-2019-bert} with 163M parameters and similarly uses WordPiece tokenization \citep{wu_2016_wordpiece}. To optimize it for the task at hand, we introduce changes to the architecture and training regimen as described below. Note that, due to time and compute constraints, all hyperparameters were optimized on the development set of Subtask 1.2 then applied as-is to the other three subtasks. We report the result of the best ensemble for each subtask.

\paragraph{General}
The experiments described below all use the following configuration: The classification head is a softmax layer over the CLS vector of the last layer $\mathbf{z}_L$ (with a dropout rate of $30\%$ during training). The base learning rate of the classification head is set higher (${1e}{-2}$) than the rest of the trainable parameters (${5e}{-6}$). The LR schedule is warmed up linearly over $250$ steps, then decayed once every 10 gradient updates following an inverse-square-root scheme to the minimum $0.01\cdot{LR}_\text{base}$. We use the Adam optimizer with decoupled weight regularization \citep{adamw}. During training, we evaluate the model every $100$ mini-batches of size $32$, and halt if the dev macro-F1 score does not improve for 10 consecutive evaluations. The maximum sequence length is $90$ for the DA dataset and $110$ for the MSA dataset.

\paragraph{Fine-tuning}
We fine-tune the full MARBERT transformer using the base configuration.

\paragraph{Adapters}
Here we embed two additional layers at each transformer block (one after the Multi-Head Attention module and one after the FFN module) following the \citet{houlsby2019parameterefficient} architecture. This allows us to preserve the pre-trained embedded knowledge in the MARBERT layers, which are trained on a rich corpus with less bias towards specific dialects. The final architecture of a transformer block is illustrated in Figure \ref{fig:transformer_archi}.

\begin{figure}[H]
    \centering
    \includegraphics[width=0.49\textwidth]{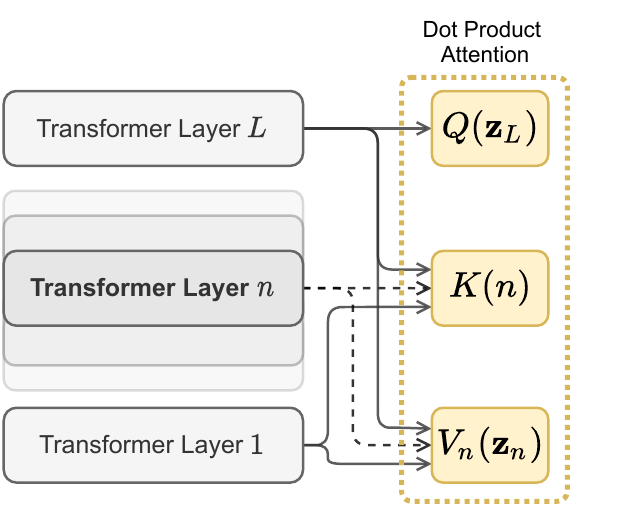}
    \caption{Vertical Attention}
    \label{fig:vatt_diagram}
\end{figure}

\paragraph{Vertical Attention (VAtt)}
The MARBERT model has $L=12$ transformer layers. For each $n \in L$, let $\mathbf{z}_n$ be the CLS token at level $n$ (after the adapter).
Let $k_n$ be a static learned positional embedding for level $n$. Apply a scaled dot-product based attention module where: query is $Q(\mathbf{z}_L)$, keys are $\{K(k_n)\}_n^L$, and values are $\left\{V_n\left(\mathbf{z}_n\right)\right\}_n^L$.
This attends over the layers' sentence representations by content-to-level (depth) addressing. We introduce this to allow choice of the abstraction level. See Figure \ref{fig:vatt_diagram}.

\paragraph{Ensembling}
We create an ensemble of multiple models on combinations of the following architectural variables:
\begin{itemize}
    \item Whether Vertical Attention is used.
    \item Training adapters or fine-tuning full model.
\end{itemize}

The soft-max outputs of the models are aggregated together by doing an element-wise multiplication. The best ensemble provide 1.13\% F1 boost over the best solo model (see Table \ref{tab:ablation_results}). 

\section{Results}
\newcommand{\textib}[1]{\textit{\textbf{#1}}}
\begin{table}[h]
\centering
\begin{tabular}{lcccc}
\hline
\multicolumn{1}{c}{\multirow{2}{*}{\textbf{Models}}} & \multicolumn{2}{c}{\textbf{DEV}} & \multicolumn{2}{c}{\textbf{TEST}} \\ \cline{2-5} 
\multicolumn{1}{c}{}                                 & \textbf{Acc.}    & \textbf{F1}   & \textbf{Acc.}    & \textbf{F1}    \\ \hline
\textbf{Adapters}                                    & \textib{52.48}   & 32.10         &   50.62          & 30.78          \\
\textbf{+VAtt}                                       & 52.28            & 31.73         &   \textib{51.08} & 30.09          \\
\textbf{Fine-tuning}                                 &  51.02           & 32.23         &   50.28          & 30.41          \\
\textbf{+VAtt}                                       &  50.07           & \textib{32.90}&   49.42          & \textib{31.30} \\ \hline
\textbf{Ensemble}                                    &  \textbf{53.42}  & \textbf{34.03}&   \textbf{51.66} & \textbf{32.26} \\ \hline
\end{tabular}
\caption{Ablation study (Subtask 1.2)}
\label{tab:ablation_results}
\end{table}




\begin{table*}[ht]
\centering
\begin{tabular}{cccccc}
\hline
\multirow{2}{*}{\textbf{Subtask}} & \multirow{2}{*}{\textbf{Models}} & \multicolumn{2}{c}{\textbf{DEV}}                    & \multicolumn{2}{c}{\textbf{TEST}} \\ \cline{3-6} 
                                  &                                  & \textbf{Acc.}  & \textbf{F1}   & \textbf{Acc.}      & \textbf{F1}                       \\ \hline
\textbf{1.1}                      & \textbf{Ours}         & \textbf{39.06} & \textbf{23.52}                     &  \textbf{35.72}           &   \textbf{22.38}      \\ \hline
  \multirow{2}{*}{\textbf{1.2}}   & MARBERT                          & 48.86          & 26.40                              & 48.40            & 29.14          \\
                                  & \textbf{Ours}         & \textbf{53.42} & \textbf{34.03}                     &   \textbf{51.66}              &  \textbf{32.26}               \\ \hline
\textbf{2.1}                      & \textbf{Ours}         & \textbf{7.04}  & \textbf{6.73}                     &   \textbf{6.66}               &   \textbf{6.43}        \\ \hline
   \multirow{2}{*}{\textbf{2.2}}  & MARBERT                          & 7.91           & 5.23                               & 8.48             & 6.28           \\
                                  & \textbf{Ours}         & \textbf{10.74} & \textbf{10.02}                      &  \textbf{9.46}           &    \textbf{8.60}      \\ \hline
\end{tabular}

\caption{Results compared to previous SOTA. MARBERT taken from \citet{mageed2020marbert}.} 
\label{tab:results}
\end{table*}

The results table shows that the best F1 score is obtained by ensembling the following list of model configurations (all of which use a maximum sequence length of $90$): (1) Fine-tuning, (2) Adapters + VAtt, (3) Fine-tuning + VAtt, and (4) Fine-tuning using a linear learning rate schedule instead of the inverse-square-root scheme. Among solo models, the best performer is the fully fine-tuned variant with Vertical Attention, with an F1 score of $32.90$ on the development set. 




\section{Discussion}

\begin{figure}[ht!]
    \centering
    \includegraphics[width=0.48\textwidth]{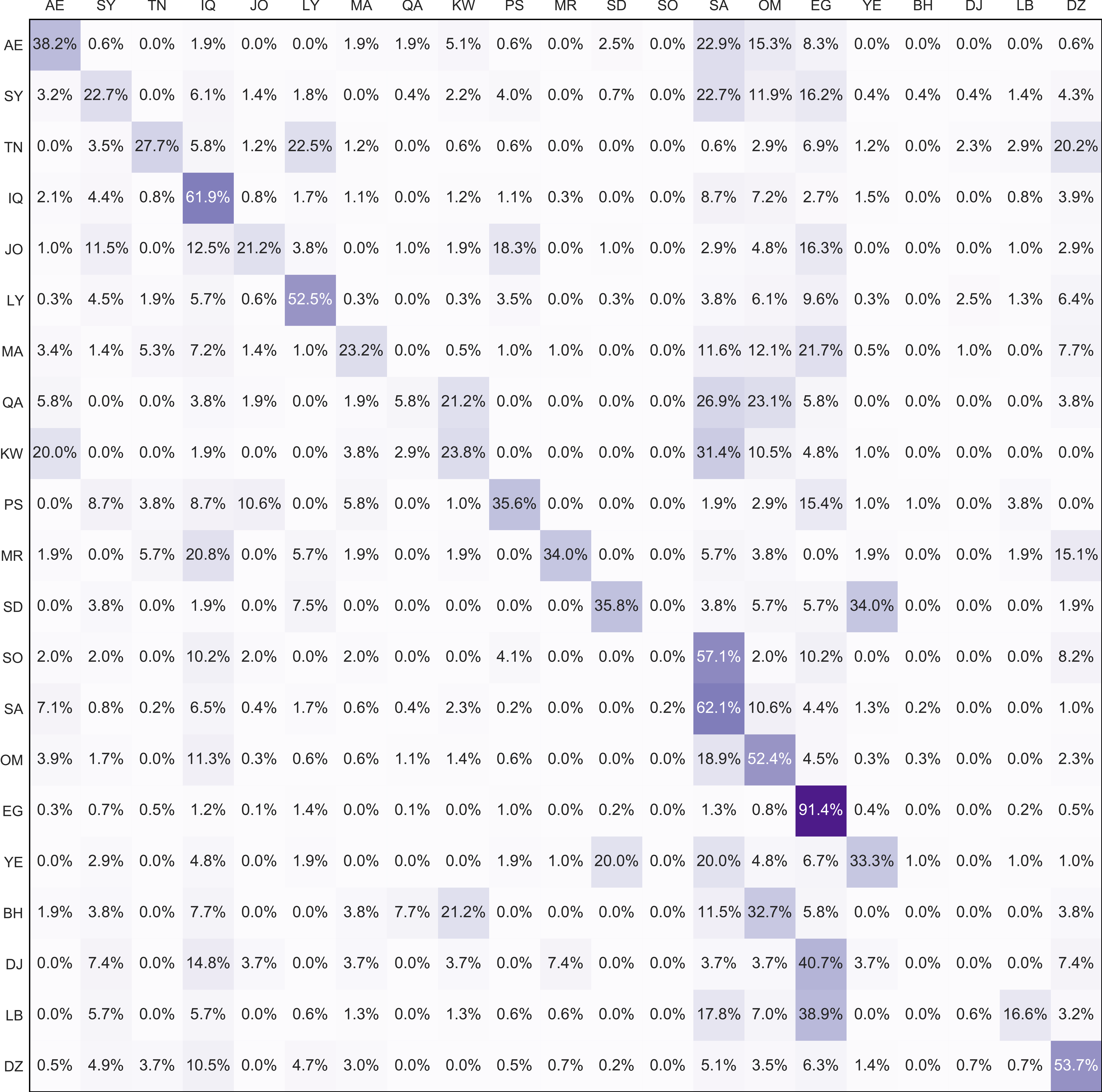}
    \caption{Confusion matrix of the predictions of the best performing model in subtask 1.2.}
    \label{fig:cm}
\end{figure}

The confusion matrix of the Ensemble in Subtask 1.2 (Table \ref{tab:ablation_results}) shows over-prediction of Egyptian and Saudi Arabian, reflecting their over-representation in training (Figure \ref{fig:train_dev_counts}). The matrix suggests that the dialects most confused together are often from geographically close countries. For example: Tunisia and Libya, and Qatar and Bahrain. Gulf countries also show high confusion.


\begin{table}[h]
\centering
\begin{tabular}{ccc}
\cline{1-3}
    & \textbf{Mean/Variance ($\lambda$)} & \textbf{Median} \\ \cline{1-3}
\textbf{Correct} & 7.16   & 5.45  \\ 
\textbf{Wrong}   & 4.22   & 3.40  \\ 
\textbf{All}     & 4.10   & 3.26  \\ \cline{1-3}
\end{tabular}
\caption{Sequence length (fitted to an Erlang distribution)}
\end{table}

Considering the sequence length in words for correct and wrong predictions, we find that correctly predicted sentences tend to be longer than the means of all and of wrongly predicted sentences (fit as an Erlang distribution). The numbers were computed after removing the special USER and URL tokens.

\begin{figure}[h]
    \centering
    \includegraphics[width=0.49\textwidth]{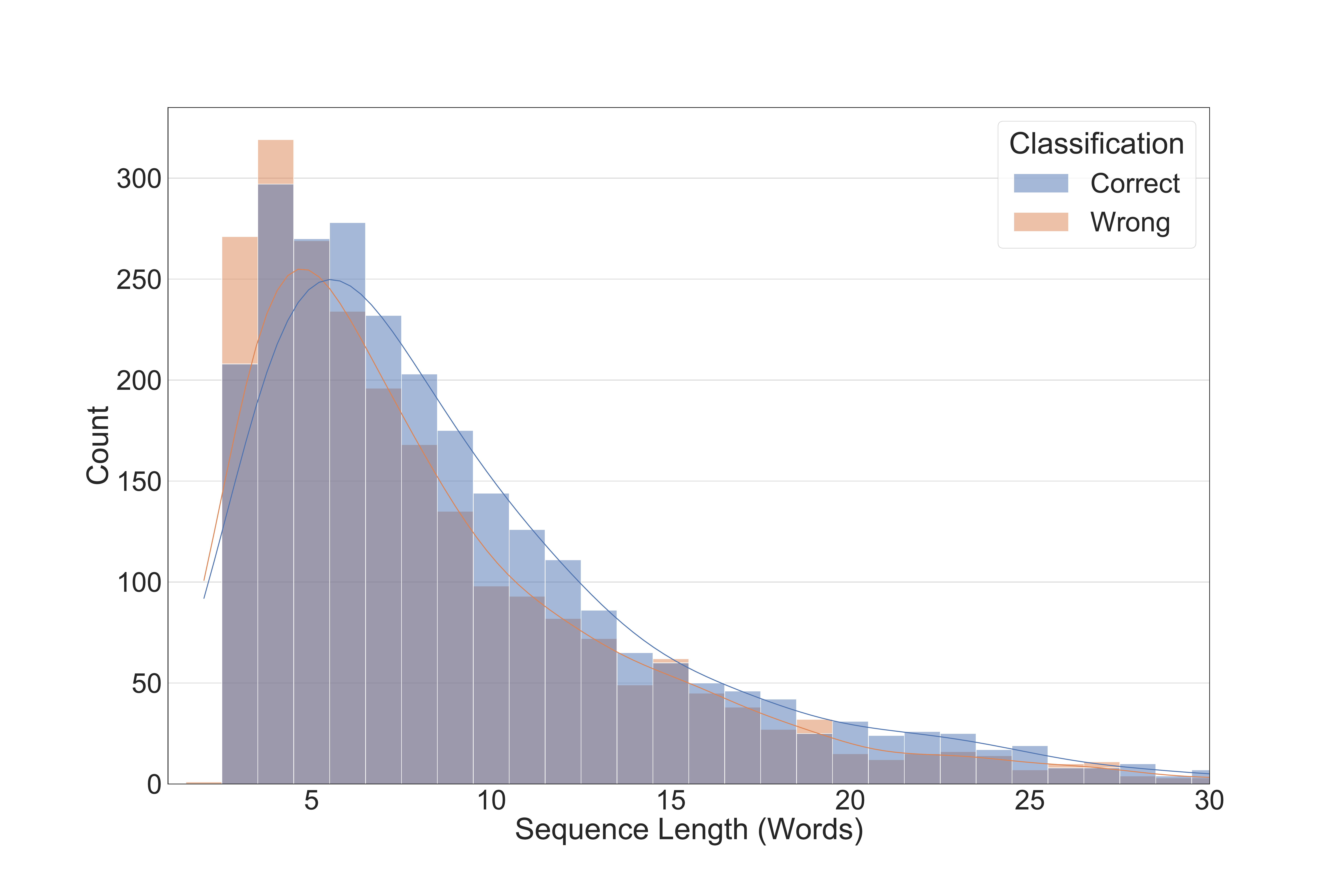}
    \caption{Distribution of length for correct and wrong classifications in subtask 1.2 trimmed at the tail after length $> 30$.}
    \label{fig:correct_wrong_lengths}
\end{figure}


\section{Conclusion}

In this paper, we present a solution that provides a new state-of-the-art on all of the NADI subtasks. Inserting adapters at each of the MARBERT transformer layers preserved the original pre-trained knowledge, stemming from a rich corpus of tweets, while still embedding task knowledge. To further improve the model's performance, we vertically attend on all of the CLS token hidden-states coming out of each of the transformer layers instead of just using the top layer's output. We ensemble models on two design variables: Whether fine-tuning is full or through adapters, and whether Vertical Attention is used. 
The bias towards dominating classes in the task dataset remains a significant issue that is still not mitigated. As a future work, we would like to employ adapter fusion \cite{pfeiffer2021adapterfusion} to attenuate this bias. 



\bibliography{anthology, paper}

\begin{thebibliography}{18}
\expandafter\ifx\csname natexlab\endcsname\relax\def\natexlab#1{#1}\fi

\bibitem[{Abdelali et~al.(2020)Abdelali, Mubarak, Samih, Hassan, and
  Darwish}]{abdelali2020arabic}
Ahmed Abdelali, Hamdy Mubarak, Younes Samih, Sabit Hassan, and Kareem Darwish.
  2020.
\newblock \href {http://arxiv.org/abs/2005.06557} {Arabic dialect
  identification in the wild}.

\bibitem[{Abdul-Mageed et~al.(2020{\natexlab{a}})Abdul-Mageed, Elmadany, and
  Nagoudi}]{mageed2020marbert}
Muhammad Abdul-Mageed, AbdelRahim Elmadany, and El~Moatez~Billah Nagoudi.
  2020{\natexlab{a}}.
\newblock Arbert \& marbert: Deep bidirectional transformers for arabic.
\newblock \emph{arXiv preprint arXiv:2101.01785}.

\bibitem[{Abdul-Mageed et~al.(2020{\natexlab{b}})Abdul-Mageed, Zhang, Bouamor,
  and Habash}]{abdulmageed2020nadi}
Muhammad Abdul-Mageed, Chiyu Zhang, Houda Bouamor, and Nizar Habash.
  2020{\natexlab{b}}.
\newblock {NADI 2020: The First Nuanced Arabic Dialect Identification Shared
  Task}.
\newblock In \emph{Proceedings of the Fifth Arabic Natural Language Processing
  Workshop (WANLP 2020)}, Barcelona, Spain.

\bibitem[{Abdul-Mageed et~al.(2021)Abdul-Mageed, Zhang, Elmadany, Bouamor, and
  Habash}]{mageed2021nadi}
Muhammad Abdul-Mageed, Chiyu Zhang, AbdelRahim Elmadany, Houda Bouamor, and
  Nizar Habash. 2021.
\newblock {NADI 2021: The Second Nuanced Arabic Dialect Identification Shared
  Task}.
\newblock In \emph{Proceedings of the Sixth {A}rabic Natural Language
  Processing Workshop (WANLP 2021)}.

\bibitem[{Abu~Farha and Magdy(2020)}]{abu-farha-magdy-2020-arabic}
Ibrahim Abu~Farha and Walid Magdy. 2020.
\newblock \href {https://www.aclweb.org/anthology/2020.osact-1.5} {From
  {A}rabic sentiment analysis to sarcasm detection: The {A}r{S}arcasm dataset}.
\newblock In \emph{Proceedings of the 4th Workshop on Open-Source Arabic
  Corpora and Processing Tools, with a Shared Task on Offensive Language
  Detection}, pages 32--39, Marseille, France. European Language Resource
  Association.

\bibitem[{Antoun et~al.(2020)Antoun, Baly, and Hajj}]{antoun2020arabert}
Wissam Antoun, Fady Baly, and Hazem Hajj. 2020.
\newblock \href {http://arxiv.org/abs/2003.00104} {Arabert: Transformer-based
  model for arabic language understanding}.

\bibitem[{Bapna and Firat(2019)}]{bapna-firat-2019-simple}
Ankur Bapna and Orhan Firat. 2019.
\newblock \href {https://doi.org/10.18653/v1/D19-1165} {Simple, scalable
  adaptation for neural machine translation}.
\newblock In \emph{Proceedings of the 2019 Conference on Empirical Methods in
  Natural Language Processing and the 9th International Joint Conference on
  Natural Language Processing (EMNLP-IJCNLP)}, pages 1538--1548, Hong Kong,
  China. Association for Computational Linguistics.

\bibitem[{Bouamor et~al.(2019)Bouamor, Hassan, and
  Habash}]{bouamor-etal-2019-madar}
Houda Bouamor, Sabit Hassan, and Nizar Habash. 2019.
\newblock \href {https://doi.org/10.18653/v1/W19-4622} {The {MADAR} shared task
  on {A}rabic fine-grained dialect identification}.
\newblock In \emph{Proceedings of the Fourth Arabic Natural Language Processing
  Workshop}, pages 199--207, Florence, Italy. Association for Computational
  Linguistics.

\bibitem[{Devlin et~al.(2019)Devlin, Chang, Lee, and
  Toutanova}]{devlin-etal-2019-bert}
Jacob Devlin, Ming-Wei Chang, Kenton Lee, and Kristina Toutanova. 2019.
\newblock \href {https://doi.org/10.18653/v1/N19-1423} {{BERT}: Pre-training of
  deep bidirectional transformers for language understanding}.
\newblock In \emph{Proceedings of the 2019 Conference of the North {A}merican
  Chapter of the Association for Computational Linguistics: Human Language
  Technologies, Volume 1 (Long and Short Papers)}, pages 4171--4186,
  Minneapolis, Minnesota. Association for Computational Linguistics.

\bibitem[{El~Mekki et~al.(2020)El~Mekki, Alami, Alami, Khoumsi, and
  Berrada}]{el-mekki-etal-2020-weighted}
Abdellah El~Mekki, Ahmed Alami, Hamza Alami, Ahmed Khoumsi, and Ismail Berrada.
  2020.
\newblock \href {https://www.aclweb.org/anthology/2020.wanlp-1.27} {Weighted
  combination of {BERT} and n-{GRAM} features for nuanced {A}rabic dialect
  identification}.
\newblock In \emph{Proceedings of the Fifth Arabic Natural Language Processing
  Workshop}, pages 268--274, Barcelona, Spain (Online). Association for
  Computational Linguistics.

\bibitem[{Gadalla(1997)}]{callhomedataset}
Hassan~and Gadalla. 1997.
\newblock Callhome egyptian arabic transcripts ldc97t19.

\bibitem[{Houlsby et~al.(2019)Houlsby, Giurgiu, Jastrzebski, Morrone,
  de~Laroussilhe, Gesmundo, Attariyan, and
  Gelly}]{houlsby2019parameterefficient}
Neil Houlsby, Andrei Giurgiu, Stanislaw Jastrzebski, Bruna Morrone, Quentin
  de~Laroussilhe, Andrea Gesmundo, Mona Attariyan, and Sylvain Gelly. 2019.
\newblock \href {http://arxiv.org/abs/1902.00751} {Parameter-efficient transfer
  learning for nlp}.

\bibitem[{Loshchilov and Hutter(2019)}]{adamw}
Ilya Loshchilov and Frank Hutter. 2019.
\newblock \href {https://openreview.net/forum?id=Bkg6RiCqY7} {Decoupled weight
  decay regularization}.
\newblock In \emph{International Conference on Learning Representations}.

\bibitem[{Pfeiffer et~al.(2021)Pfeiffer, Kamath, Rücklé, Cho, and
  Gurevych}]{pfeiffer2021adapterfusion}
Jonas Pfeiffer, Aishwarya Kamath, Andreas Rücklé, Kyunghyun Cho, and Iryna
  Gurevych. 2021.
\newblock \href {http://arxiv.org/abs/2005.00247} {Adapterfusion:
  Non-destructive task composition for transfer learning}.

\bibitem[{Pfeiffer et~al.(2020)Pfeiffer, R\"uckl\'{e}, Poth, Kamath, Vuli\'{c},
  Ruder, Cho, and Gurevych}]{pfeiffer2020AdapterHub}
Jonas Pfeiffer, Andreas R\"uckl\'{e}, Clifton Poth, Aishwarya Kamath, Ivan
  Vuli\'{c}, Sebastian Ruder, Kyunghyun Cho, and Iryna Gurevych. 2020.
\newblock \href {https://www.aclweb.org/anthology/2020.emnlp-demos.7}
  {{AdapterHub: A Framework for Adapting Transformers}}.
\newblock In \emph{Proceedings of the 2020 Conference on Empirical Methods in
  Natural Language Processing (EMNLP 2020): Systems Demonstrations}, pages
  46--54, Online. Association for Computational Linguistics.

\bibitem[{Talafha et~al.(2020)Talafha, Ali, Za'ter, Seelawi, Tuffaha, Samir,
  Farhan, and Al-Natsheh}]{talafha2020multidialect}
Bashar Talafha, Mohammad Ali, Muhy~Eddin Za'ter, Haitham Seelawi, Ibraheem
  Tuffaha, Mostafa Samir, Wael Farhan, and Hussein~T. Al-Natsheh. 2020.
\newblock \href {http://arxiv.org/abs/2007.05612} {Multi-dialect arabic bert
  for country-level dialect identification}.

\bibitem[{Wu et~al.(2016)Wu, Schuster, Chen, Le, Norouzi, Macherey, Krikun,
  Cao, Gao, Macherey, Klingner, Shah, Johnson, Liu, Kaiser, Gouws, Kato, Kudo,
  Kazawa, and Dean}]{wu_2016_wordpiece}
Yonghui Wu, Mike Schuster, Zhifeng Chen, Quoc Le, Mohammad Norouzi, Wolfgang
  Macherey, Maxim Krikun, Yuan Cao, Qin Gao, Klaus Macherey, Jeff Klingner,
  Apurva Shah, Melvin Johnson, Xiaobing Liu, ukasz Kaiser, Stephan Gouws,
  Yoshikiyo Kato, Taku Kudo, Hideto Kazawa, and Jeffrey Dean. 2016.
\newblock Google's neural machine translation system: Bridging the gap between
  human and machine translation.

\bibitem[{Zaidan and Callison-Burch(2014)}]{AOC:1}
Omar~F. Zaidan and Chris Callison-Burch. 2014.
\newblock \href {https://doi.org/10.1162/COLI\_a\_00169} {Arabic dialect
  identification}.
\newblock \emph{Computational Linguistics}, 40(1):171--202.

\end{thebibliography}
\bibliographystyle{acl_natbib}

\appendix

\section{Appendices}
\label{sec:appendix}


\subsection{System Architecture}
\begin{figure}[h]
    \centering
    \includegraphics[width=0.24\textwidth]{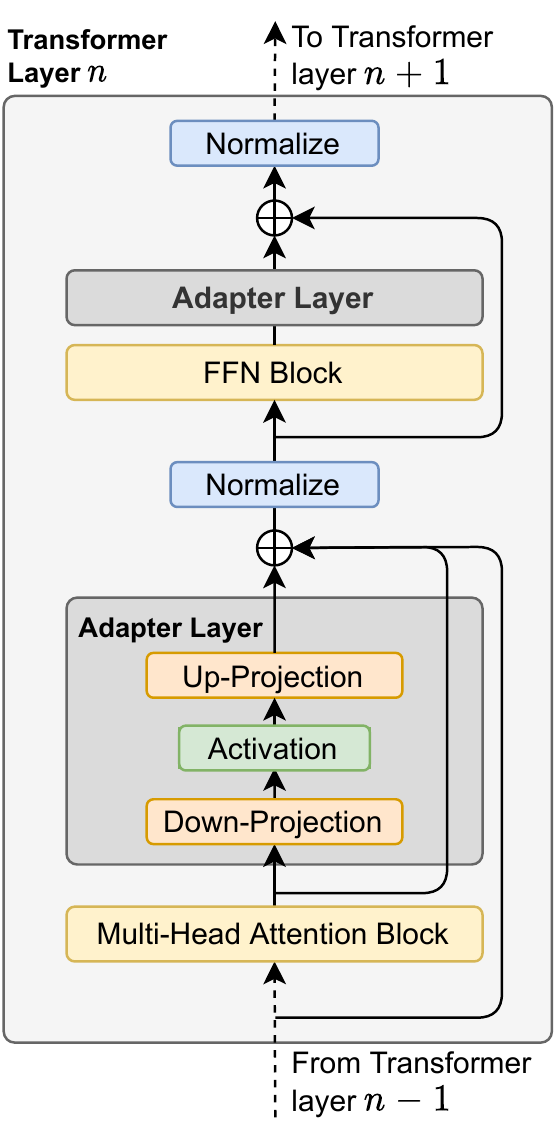}
    \caption{Details of the used model's architecture, specifically looking at one transformer layer. All layers are of the same architecture.\footnotemark}
    \label{fig:transformer_archi}
\end{figure}
\footnotetext{Diagrams generated using \url{diagrams.net} (draw.io).}

\end{document}